\documentclass[a4paper, 11pt]{article}
\usepackage[utf8]{inputenc}
\usepackage[margin = 2.5cm]{geometry}

\usepackage{amssymb}
\usepackage{latexsym}

\usepackage{url}
\usepackage{xcolor}
\usepackage{multirow}
\usepackage{epsfig,subfigure}
\usepackage{pgfplots}
\usepackage{dblfloatfix}
\usepackage{adjustbox}
\definecolor{newcolor}{rgb}{.8,.349,.1}

\title{Semi-supervised deep learning based on label propagation in a 2D embedded space}

\author{Bárbara C. Benato\footnote{barbara.benato@ic.unicamp.br}, Jancarlo F. Gomes, Alexandru C. Telea, Alexandre X. Falcão\\ University of Campinas, Campinas, Brazil} 

\date{}

\begin{document}

%\begin{frontmatter}

%\renewcommand{\baselinestretch}{0.9} 

\maketitle

%200 words
\begin{abstract}
While convolutional neural networks need large labeled sets for training images, expert human supervision of such datasets can be very laborious. Proposed solutions propagate labels from a small set of supervised images to a large set of unsupervised ones to obtain sufficient truly-and-artificially labeled samples to train a deep neural network model. Yet, such solutions need many supervised images for validation. We present a loop in which a deep neural network (VGG-16) is trained from a set with more correctly labeled samples along iterations, created by using t-SNE to project the features of its last max-pooling layer into a 2D embedded space in which labels are propagated using the Optimum-Path Forest semi-supervised classifier. As the labeled set improves along iterations, it improves the features of the neural network. We show that this can significantly improve classification results on test data (using only 1\% to 5\% of supervised samples) of three private challenging datasets and two public ones. 
\end{abstract}

%\begin{keyword} %six
%\MSC 41A05\sep 41A10\sep 65D05\sep 65D17

%\end{frontmatter}

%\linenumbers

%% main text

\section{Introduction}

Convolutional neural networks (CNNs) usually need large training sets (labeled images)~\cite{Lin:2014,imagenet:2015}. While solutions to deal with this problem may use regularization, fine-tuning, transfer learning, and data augmentation~\cite{Sun:2017}, manually annotating a sufficient number of images (human supervision) remains a great obstacle, notably when expert users are needed, as in Biology and Medicine. 

To build a sufficiently large training set, an interesting and under-exploited alternative is to propagate labels from a small set of supervised images to a much larger set of unsupervised ones, as shown by Lee~\cite{Lee:2013}, as an alternative for entropy regularization. Lee trained a neural network with 100 to 3000 supervised images, assigned the class with maximum predicted probability to the remaining unsupervised ones, and then fine-tuned a neural network with those truly-and-artificially labeled (\emph{pseudo} labeled) samples, showing advantages over other semi-supervised learning methods. Still, this method requires a validation set with 1000 or more supervised images for the optimization of hyper-parameters; used a network with just a single hidden layer; and was demonstrated on one dataset only (MNIST).

Label propagation from supervised to unsupervised samples was recently used to build larger training sets~\cite{Amorim:2019,Gong:2012,Iscen:2019,Wu:2018,Zhun:2018}. Amorim et al.\cite{Amorim:2019} used the semi-supervised Optimum Path Forest (OPFSemi) classifier~\cite{Amorim:2016} to do this, showing advantages for training CNNs, by outperforming several semi-supervised techniques in the literature. Yet, they did not explore CNNs pre-trained with large supervised datasets for transfer learning, and  still used many supervised samples ($10\%$ of the dataset) for validation.

Graph-based semi-supervised learning has recently received increasing attention~\cite{Amorim:2014,Amorim:2016,Belkin:2004,Chapelle:2009,Zhu:2005,Zhu:2008}. By interpreting training samples as nodes of a graph, whose arcs connect adjacent samples in the feature space, one can exploit adjacency relations and connectivity functions to propagate labels from supervised samples to their most strongly connected unsupervised ones. Benato et al.~\cite{BenatoSibgrapi:2018} showed the advantages of OPFSemi for label propagation in a 2D embedded space created by t-SNE~\cite{MaatenJMLR:14} from the latent space of an auto-encoder trained with unsupervised images. They showed that supervised classifiers can achieve higher performance on unseen test sets, when trained with larger sets of truly-and-artificially labeled samples, and that OPFSemi surpassed LapSVM~\cite{Sindhwani:2005} for label propagation. Yet, they have not used this strategy to train deep neural networks. 

We fill the gap in previous work by proposing a loop (Fig.~\ref{f.pipeline}) that trains a deep neural network (VGG-16,~\cite{VGG16}) with truly-and-artificially labeled samples along iterations. At each iteration, it generates a 2D embedded space by t-SNE projection of the  features at its last max-pooling layer (i.e., before the MLP) and propagates labels by OPFSemi, such that the labeled set jointly improves with the feature space of the CNN over iterations. We demonstrate this loop can improve classification on unseen test data of challenging datasets.

\section{Proposed Pipeline}
\begin{figure}[!t]
\centering
\includegraphics[width=1.0\linewidth]{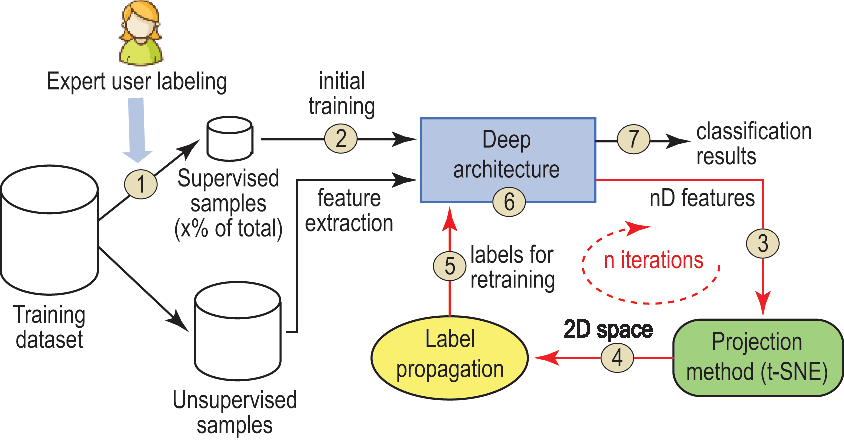}
\caption{Pipeline of our method. The user supervises a small percentage $x$ of images (1). These images are used to train a deep neural network (2), which extracts features from the unsupervised images (3). These features are projected in a 2D embedded space (4). A semi-supervised classifier propagates labels to the unsupervised images (5). Then, the model is retrained by all images and their assigned labels (6), creating a new and improved feature space along iterations. Finally, classification results are obtained from the neural network model (7).}
\label{f.pipeline}
\end{figure}

After the user supervises a small percentage $x$ of training images, our method executes a loop with three steps, \emph{deep feature learning}, \emph{feature space projection}, and \emph{label propagation} (Fig.~\ref{f.pipeline}), as follows.

%\noindent\emph{Deep feature learning}: Due to the ability of pre-trained deep neural networks to transfer knowledge between scenarios, a common CNN architecture -- VGG-16~(\cite{VGG16}) --  is first fine-tuned with the supervised images and, in the subsequent iterations, it is trained with all truly-and-artificially labeled images.

%\noindent\emph{Feature space projection}: A 2D embedded space is generated by t-SNE projection~(\cite{MaatenJMLR:14}) from the nD features of the last hidden layer of VGG-16. Such strategy stems from the observation that semi-supervised label propagation is better in the 2D space~(\cite{BenatoSibgrapi:2018}).

%\noindent\emph{Label propagation}: A semi-supervised Optimum Path Forest classifier~(\cite{Amorim:2016}), named OPFSemi, propagates labels from the supervised to the unsupervised samples in the embedded space. The estimated labels are used to feed the deep architecture over a few iterations to obtain a new (and better) feature space for final classification.\\

\subsection{Deep Feature Learning}

As we aim a data annotation that requires minimum user effort, we consider exploring the ability of pre-trained CNNs to transfer knowledge~\cite{Yosinski:2014} between scenarios -- e.g., from natural images to medical images -- using few supervised samples and few epochs. Thus, we employ VGG-16, as pre-trained on ImageNet~\cite{imagenet:2015}, and first fine tuned it with the supervised images. In the subsequent iterations of the proposed loop, it is trained with all truly-and-artificially labeled images.  

\subsection{Feature Space Projection}

The features of the last max-pooling layer of VGG-16 are  projected by t-SNE~\cite{MaatenJMLR:14} in a 2D embedded space. One may conceptually divide a deep neural network into (a) layers for feature extraction, (b) fully connected layers for feature space reduction, (c) and the decision layer, being (b) and (c) a MLP classifier. We then explored features that result from (a), where the feature space is still high and sparse, but a comparison with the output of the last hidden layer will be done in the future. ~\cite{RauberInfVis2017} showed that high classification accuracies relate to a good separation among classes in a 2D projected space. This finding tells that if a projection can present a good separation among classes in 2D, then a good separation among classes can also be found in the nD space. So why do not use a good 2D projected space?

Benato et al.\cite{BenatoSibgrapi:2018} investigated the impact of using the 2D projected space for two distinct semi-supervised classifiers in the label propagation task. They showed that the label propagation in such embedded  space leads to better classification results when compared with the nD latent feature space of an auto-encoder.
In this way, we opt to investigate label propagation using the 2D projected space to create larger training sets for deep learning. For this, we explore the same projection method used in both studies: the t-SNE algorithm~\cite{MaatenJMLR:14}. 

\subsection{Label Propagation}
OPFSemi was used in both spaces, the 2D space generated by t-SNE\,\cite{BenatoSibgrapi:2018} and the original feature space\,\cite{Amorim:2019}. OPFSemi considers each sample as a node of a complete graph, weighted by the Euclidean distance between samples, and defines the cost of a path connecting two nodes as the maximum arc weight along it. From the training nodes, the supervised ones are used as seeds to compute a minimum-cost path forest, such that each seed assigns its label to the most closely connected unsupervised nodes of its tree. 

\section{Experiments and Results}
In this section, we describe the experimental set-up, datasets, implementation details, and results of our method.

\subsection{Experimental Set-up}
\label{ssec.setup}
First, we randomly divide each dataset into three subsets: $S$ of supervised training samples, $U$ of unsupervised training samples, and $T$ of testing samples. As we intend to notice the impact of annotated samples for classification, we let $S\cup U$ has $70\%$ of samples, while $T$ has $30\%$. To minimize the user effort, i.e., considering that the user has the effort to supervise some samples, $|S|$ has to be much smaller than $|U|$. Then we propose experiments with values of $|S|$ from $1\%$ to $5\%$ of the entire data set. For statistics, we generate three partitions of each experiment randomly and in a stratified manner. 

To validate our proposed method of Deep Feature Annotation (\emph{DeepFA looping}), we consider three experiments described next:
\begin{enumerate}
    \item \emph{baseline}: training VGG-16 on $S$, testing on $T$, ignoring $U$ (steps 1,2,6,7 in Fig.~\ref{f.pipeline}).
    \item \emph{DeepFA}: training VGG-16 on $S$; extracting $S\cup U$ features from VGG-16 and projecting them in $2D$ with t-SNE; OPFSemi label estimation in $U$; training VGG-16 on $S\cup U$ and testing on $T$ (all of Fig.~\ref{f.pipeline} for $n=1$).
    \item \emph{DeepFA-looping}: training VGG-16 on $S$; extracting $S\cup U$ features from VGG-16; projecting data in $2D$ with t-SNE; OPFSemi label propagation on $U$; training VGG-16 on $S\cup U$; repeat from the projection step $n=5$ times; testing on $T$ (all of Fig.~\ref{f.pipeline} for $n>1$).
\end{enumerate}

%\begin{figure}[!h]
%shown in Figure~\ref{f.methods},
%\centering
%\includegraphics[width=1.0\linewidth]{figs/2020PRL_methods.png}
%\caption{Proposed experiments. a) \emph{Baseline} experiment (red box) that includes VGG-16. b) \emph{DeepFA} experiment (blue box) with $n=1$ iterations, and \emph{DeepFA looping} experiment, when $n>1$.
%ALEX: I really wonder if this figure is useful, since Fig. 1 is basically giving all this information. We could describe Baseline as steps (1,2,7) in Fig. 1; DeepFA as all Fig 1, but with n=1; and DeepFA looping as all Fig 1, with $n > 1$.
%}
%\label{f.methods}
%\end{figure}

We use the final probability of VGG-16 as our metric for effectiveness comparison. From that probability, we compute the accuracy commonly used in those purposes. As we explore unbalanced datasets, we also compute Cohen's $\kappa$ coefficient, $\kappa \in [-1,1]$, where $\kappa \leq 0$ means no possibility and $\kappa = 1$ means full possibility of agreement occurring by chance, respectively. We also compute the label propagation accuracy in $U$ for the experiments in which OPFSemi propagates sample labels, defined as the number of correct labels assigned in $U$ over the number of samples in $U$. 

\subsection{Datasets}
We first consider two public datasets: MNIST\,\cite{Lecun:2010:mnist} contains  handwritten digits from $0$ to $9$ as $28\times28$ grayscale images. We used a  random subset of $5$K samples from MNIST's total of $60$K. CIFAR-10\,\cite{Cifar10} contains color images ($32\times32$ pixels) in 10 classes: airplane, automobile, bird, cat, deer, dog, frog, horse, ship, and truck. We used a random subset of $5$K images from CIFAR-10's total of $60$K images. 

\begin{figure}[!h]
\centering
\includegraphics[width=1.0\linewidth]{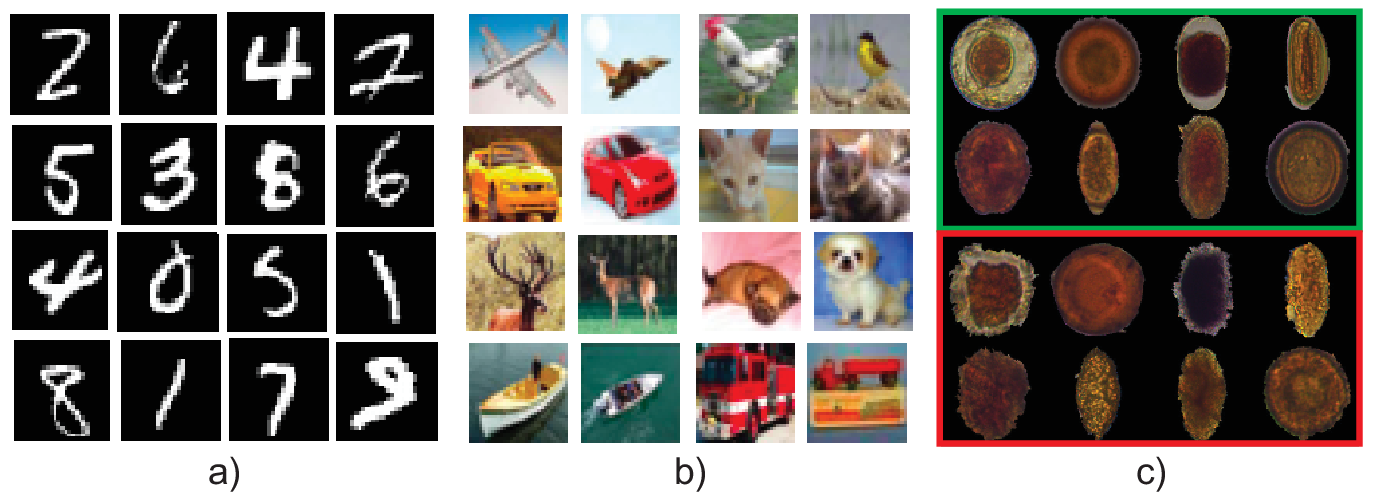}
\caption{Datasets: (a) MNIST (b) CIFAR-10 and (c) H.eggs, with parasites (green box) and similar impurities (red box).}
\label{f.datasets}
\end{figure}

We also used three private datasets from a real-world problem (see Fig.~\ref{f.datasets}). These datasets contain color microscopy images ($200\times200$ pixels), obtained from an automatic process, with the most common species of human intestinal parasites in Brazil, responsible for public health problems in most tropical countries\,\cite{Suzuki:2013}. These three datasets are challenging, since they are unbalanced and contain an impurity class with the large majority of the samples: a diverse class with samples very similar to parasites, further increasing the difficulty of the problem (see Fig.~\ref{f.datasets}). In detail: The \emph{Helminth larvae} dataset presents larvae and impurities ($2$ classes, $3514$ images); the \emph{Helminth eggs} dataset has the following categories: \emph{H.nana}, \emph{H.diminuta}, \emph{Ancilostomideo}, \emph{E.vermicularis}, \emph{A.lumbricoides}, \emph{T.trichiura}, \emph{S.mansoni}, \emph{Taenia}, and impurities ($9$ classes, $5112$ images); and the \emph{Protozoan cysts} dataset having the categories \emph{E.coli}, \emph{E.histolytica}, \emph{E.nana}, \emph{Giardia}, \emph{I.butschlii}, \emph{B.hominis}, and impurities ($7$ classes, $9568$ images). 

Recent studies tried to improve the classification results on these parasite datasets using a variety of methods --  handcrafted features~\cite{RauberInfVis2017,Suzuki:2013}, semi-supervised learning~\cite{Amorim:2016, BenatoSibgrapi:2018}, active learning~\cite{Saito:2015}, and deep learning ~\cite{peixinho,Peixinho:2018}. The deep learning approach obtained the best classification accuracy for the parasite datasets by using data augmentation, feature extraction, and finally Support Vector Machines~\cite{Hearst:1998} to classify the feature vectors. \emph{H. larvae} obtained $95.70\%$ accuracy, \emph{H. eggs} $96.79\%$, and \emph{P. cysts} $96.49\%$ respectively, when the impurities were considered. Table~\ref{t.splits} presents the experimental set-up described in Sec.~\ref{ssec.setup} for these 7 datasets.

\begin{table}[h]
\centering
\scriptsize
\caption{Number of samples in each set $S$, $U$, and $T$ considering $|S|$ for five sample percentages $x=1,2,\ldots,5\%$ of supervised images in each dataset.}
\begin{tabular}{c|c|r|r|r|r}
\textbf{Dataset} & $x$  & \multicolumn{1}{c|}{$\mid S\mid$} & \multicolumn{1}{c|}{$\mid U\mid$} & \multicolumn{1}{c|}{$\mid T\mid$} & \multicolumn{1}{c}{\textbf{Total}} \\ \hline
\multirow{5}{*}{H.larvae}
 & $1\%$ & $25$ & $1434$ & $1055$ & $2514$ \\  
 & $2\%$ & $50$ & $1409$ & $1055$ & $2514$ \\  
 & $3\%$ & $75$ & $1384$ & $1055$ & $2514$ \\  
 & $4\%$ & $100$ & $1359$ & $1055$ & $2514$ \\ 
 & $5\%$ & $125$ & $1334$ & $1055$ & $2514$ \\ \hline
\multirow{5}{*}{H.eggs}
 & $1\%$ & $17$ & $1220$ & $531$ & $1768$ \\
 & $2\%$ & $35$ & $1202$ & $531$ & $1768$ \\  
 & $3\%$ & $53$ & $1184$ & $531$ & $1768$ \\ 
 & $4\%$ & $70$ & $1167$ & $531$ & $1768$ \\ 
 & $5\%$ & $88$ & $1149$ & $531$ & $1768$ \\ \hline
\multirow{5}{*}{P.cysts}
 & $1\%$ & $38$ & $1118$ & $2696$ & $3852$ \\ 
 & $2\%$ & $77$ & $1079$ & $2696$ & $3852$ \\ 
 & $3\%$ & $115$ & $1041$ & $2696$ & $3852$ \\ 
 & $4\%$ & $154$ & $1002$ & $2696$ & $3852$ \\
 & $5\%$ & $192$ & $964$ & $2696$ & $3852$ \\ \hline
\multirow{5}{*}{H.eggs imp}
 & $1\%$ & $51$ & $3527$ & $1534$ & $5112$ \\ 
 & $2\%$ & $102$ & $3476$ & $1534$ & $5112$ \\
 & $3\%$ & $153$ & $3425$ & $1534$ & $5112$ \\
 & $4\%$ & $204$ & $3374$ & $1534$ & $5112$ \\
 & $5\%$ & $255$ & $3323$ & $1534$ & $5112$ \\ \hline
\multirow{5}{*}{P.cysts imp}
 & $1\%$ & $95$ & $6602$ & $2871$ & $9568$ \\
 & $2\%$ & $191$ & $6506$ & $2871$ & $9568$ \\
 & $3\%$ & $287$ & $6410$ & $2871$ & $9568$ \\
 & $4\%$ & $382$ & $6315$ & $2871$ & $9568$ \\
 & $5\%$ & $478$ & $6219$ & $2871$ & $9568$ \\ \hline
\multirow{5}{*}{\shortstack{MNIST /\\ CIFAR-10}}
 & $1\%$ & $50$ & $3450$ & $1500$ & $5000$ \\ 
 & $2\%$ & $100$ & $3400$ & $1500$ & $5000$ \\ 
 & $3\%$ & $150$ & $3350$ & $1500$ & $5000$ \\  
 & $4\%$ & $200$ & $3300$ & $1500$ & $5000$ \\ 
 & $5\%$ & $250$ & $3250$ & $1500$ & $5000$ \\ \hline
\end{tabular}

\label{t.splits}
\end{table}

\subsection{Implementation details}
We implemented VGG-16 in Python using Keras~\cite{Chollet:2015:keras}. We load the pre-trained weights from the ImageNet~\cite{imagenet:2015} and then fine-tuned this model using the supervised $S$ first, and subsequently  labeled sets ($S \cup U$) for each chosen dataset. To guarantee convergence, we used $100$ epochs with stochastic gradient descent with a linearly decaying learning-rate from $10^{-4}$ to zero over 100 epochs and momentum of 0.9. 

%\begin{figure}[!h]
%\centering
%    \includegraphics[width=\linewidth]{figs/learningrate.png}
%\caption{Linear learning rate decay.}
%\label{f.learningrate}
%\end{figure}

\subsection{Experimental Results}
We next use our experimental data to address three joint questions: How do (Q1) an increased number of supervised samples; (Q2) OPFSemi's labeled samples; and (Q3) iterative OPFSemi labeling increase VGG-16's effectiveness?\\

\noindent\emph{Q1: Supervised training effect:} Table~\ref{t.baseline}(a) shows mean and standard deviation for accuracy and Cohen's $\kappa$ for VGG-16 trained from $S$ with $1\% .. 5\%$ of supervised samples and tested in $T$ (\emph{baseline} experiment) for all datasets. We see an accuracy and $\kappa$ increase with the  number of supervised samples. For the H.larvae and H.eggs datasets, this trend can not be seen  in accuracy and $\kappa$ for given training-data percentages ($3\%$ to $4\%$). Still, even for small training-data amounts, VGG-16 performs better when the number of supervised training samples increases.\\

\begin{table*}[!tbh]
\scriptsize
\centering

\caption{\label{t.baseline} Results of the \emph{baseline}, \emph{DeepFA}, and \emph{DeepFA looping} experiments, all datasets, five supervised sample percentages $x$. The best results for each dataset and percentages $x$ are in bold.}
\begin{adjustbox}{width=0.9\textwidth}
\begin{tabular}{|l|l|l|c|c|c|c|c|}
\hline
& \textbf{Dataset} & \textbf{Metrics} & $x=1\%$ & $x=2\%$ & $x=3\%$ & $x=4\%$ & $x=5\%$ \\ \hline
\multirow{14}{*}{\rotatebox[origin=c]{90}{\textbf{(a) Baseline experiment}}} &
\multicolumn{1}{l|}{\multirow{2}{*}{H.larvae}} & accuracy \hspace{0.25cm} & 0.930806 $\pm$ 0.0266 & 0.958925 $\pm$ 0.0038 & 0.961138 $\pm$ 0.0066 & 0.960821 $\pm$ 0.0070 & 0.971564 $\pm$ 0.0068 \\
& \multicolumn{1}{l|}{} & kappa & 0.613432 $\pm$ 0.2334 & 0.824394 $\pm$ 0.0280 & 0.818082 $\pm$ 0.0492 & 0.808397 $\pm$ 0.0416 & 0.868460 $\pm$ 0.0361 \\ 
\cline{2-8}
& \multicolumn{1}{l|}{\multirow{2}{*}{H.eggs}} & accuracy & 0.812932 $\pm$ 0.0599 & 0.925926 $\pm$ 0.0189 & 0.934714 $\pm$ 0.0313 & 0.929065 $\pm$ 0.0132 & 0.966101 $\pm$ 0.0032 \\
& \multicolumn{1}{l|}{} & kappa & 0.775954 $\pm$ 0.0737 & 0.912299 $\pm$ 0.0224 & 0.923002 $\pm$ 0.0367 & 0.916335 $\pm$ 0.0154 & 0.959807 $\pm$ 0.0039 \\ 
\cline{2-8}
& \multicolumn{1}{l|}{\multirow{2}{*}{P.cysts}} & accuracy & 0.757209 $\pm$ 0.0158 & 0.881776 $\pm$ 0.0113 & 0.913783 $\pm$ 0.0079 & 0.914648 $\pm$ 0.0077 & 0.934545 $\pm$ 0.0133 \\
& \multicolumn{1}{l|}{} & kappa & 0.651933 $\pm$ 0.0232 & 0.837102 $\pm$ 0.0163 & 0.882551 $\pm$ 0.0108 & 0.884303 $\pm$ 0.0108 & 0.912294 $\pm$ 0.0177 \\ 
\cline{2-8}
& \multicolumn{1}{l|}{\multirow{2}{*}{H.eggs imp}} & accuracy & 0.862234 $\pm$ 0.0157 & 0.900696 $\pm$ 0.0087 & 0.910256 $\pm$ 0.0167 & 0.931986 $\pm$ 0.0057 & 0.937419 $\pm$ 0.0086 \\
& \multicolumn{1}{l|}{} & kappa & 0.740861 $\pm$ 0.0287 & 0.815160 $\pm$ 0.0138 & 0.833168 $\pm$ 0.0301 & 0.876969 $\pm$ 0.0090 & 0.886231 $\pm$ 0.0159 \\ 
\cline{2-8}
& \multicolumn{1}{l|}{\multirow{2}{*}{P.cysts imp}} & accuracy & 0.850691 $\pm$ 0.0189 & 0.865320 $\pm$ 0.0018 & 0.900383 $\pm$ 0.0072 & 0.903634 $\pm$ 0.0129 & 0.916173 $\pm$ 0.0045 \\
& \multicolumn{1}{l|}{} & kappa & 0.751667 $\pm$ 0.0280 & 0.776938 $\pm$ 0.0031 & 0.832300 $\pm$ 0.0106 & 0.840126 $\pm$ 0.0216 & 0.860640 $\pm$ 0.0076 \\ 
\cline{2-8}
& \multicolumn{1}{l|}{\multirow{2}{*}{MNIST}} & accuracy & 0.661111 $\pm$ 0.0523 & 0.782222 $\pm$ 0.0269 & 0.870445 $\pm$ 0.0050 & 0.876444 $\pm$ 0.0132 & 0.909778 $\pm$ 0.0143 \\
& \multicolumn{1}{l|}{} & kappa & 0.623148 $\pm$ 0.0582 & 0.757848 $\pm$ 0.0298 & 0.855944 $\pm$ 0.0056 & 0.862635 $\pm$ 0.0147 & 0.899686 $\pm$ 0.0159 \\ 
\cline{2-8}
& \multicolumn{1}{l|}{\multirow{2}{*}{CIFAR10}} & accuracy & 0.266000 $\pm$ 0.0264 & 0.321555 $\pm$ 0.0151 & 0.372889 $\pm$ 0.0341 & 0.417111 $\pm$ 0.0413 & 0.455333 $\pm$ 0.0263 \\
& \multicolumn{1}{l|}{} & kappa & 0.183681 $\pm$ 0.0301 & 0.245770 $\pm$ 0.0166 & 0.303050 $\pm$ 0.0377 & 0.352095 $\pm$ 0.0461 & 0.394558 $\pm$ 0.0291 \\ \hline
\end{tabular}
\end{adjustbox}
\begin{adjustbox}{width=0.9\textwidth}
\begin{tabular}{|c|l|l|c|c|c|c|c|}
\hline
\multirow{21}{*}{\rotatebox[origin=c]{90}{\textbf{(b) DeepFA experiment}}} &
\multicolumn{1}{l|}{\multirow{3}{*}{H.larvae}} & accuracy & \textbf{0.962085} $\pm$ \textbf{0.0148} & 0.968721 $\pm$ 0.0053 & 0.967773 $\pm$ 0.0047 & 0.974092 $\pm$ 0.0112 & 0.977567 $\pm$ 0.0043 \\
& \multicolumn{1}{l|}{} & kappa & \textbf{0.819799} $\pm$ \textbf{0.0767} & 0.864692 $\pm$ 0.0251 & 0.854607 $\pm$ 0.0308 & 0.878935 $\pm$ 0.0624 & 0.900290 $\pm$ 0.0197 \\ 
& \multicolumn{1}{l|}{} & propagation & \textbf{0.961366} $\pm$ \textbf{0.0132} & 0.969364 $\pm$ 0.0070 & 0.963806 $\pm$ 0.0108 & 0.975193 $\pm$ 0.0135 & 0.979531 $\pm$ 0.0048 \\ 
\cline{2-8}
& \multicolumn{1}{l|}{\multirow{3}{*}{H.eggs}} & accuracy & 0.949780 $\pm$ 0.0104 & 0.967985 $\pm$ 0.0082 & 0.962963 $\pm$ 0.0120 & \textbf{0.974263} $\pm$ \textbf{0.0203} & 0.978656 $\pm$ 0.0141 \\
& \multicolumn{1}{l|}{} & kappa & 0.940671 $\pm$ 0.0122 & 0.962146 $\pm$ 0.0097 & 0.956316 $\pm$ 0.0234 & \textbf{0.969631} $\pm$ \textbf{0.0239} & 0.974742 $\pm$ 0.0167 \\ 
& \multicolumn{1}{l|}{} & propagation & 0.957154 $\pm$ 0.0127 & 0.974400 $\pm$ 0.0055 & 0.971436 $\pm$ 0.0084 & \textbf{0.978442} $\pm$ \textbf{0.0102} & 0.987065 $\pm$ 0.0037 \\ 
\cline{2-8}
& \multicolumn{1}{l|}{\multirow{3}{*}{P.cysts}} & accuracy & 0.847751 $\pm$ 0.0106 & 0.912341 $\pm$ 0.0154 & 0.914072 $\pm$ 0.0269 & 0.937428 $\pm$ 0.0018 & 0.950404 $\pm$ 0.0178 \\
& \multicolumn{1}{l|}{} & kappa & 0.794713 $\pm$ 0.0124 & 0.882537 $\pm$ 0.0213 & 0.885475 $\pm$ 0.0348 & 0.916035 $\pm$ 0.0028 & 0.933649 $\pm$ 0.0235 \\
& \multicolumn{1}{l|}{} & propagation & 0.832715 $\pm$ 0.0153 & 0.897997 $\pm$ 0.0231 & 0.893793 $\pm$ 0.0303 & 0.931627 $\pm$ 0.0042 & 0.937685 $\pm$ 0.0161 \\ 
\cline{2-8}
& \multicolumn{1}{l|}{\multirow{3}{*}{H.eggs imp}} & accuracy & 0.928509 $\pm$ 0.0032 & 0.941113 $\pm$ 0.0010 & 0.935246 $\pm$ 0.0053 & 0.948501 $\pm$ 0.0111 & \textbf{0.956758} $\pm$ \textbf{0.0046} \\
& \multicolumn{1}{l|}{} & kappa & 0.873674 $\pm$ 0.0068 & 0.895627 $\pm$ 0.0014 & 0.885487 $\pm$ 0.0104 & 0.908733 $\pm$ 0.0190 & \textbf{0.923366} $\pm$ \textbf{0.0079} \\ 
& \multicolumn{1}{l|}{} & propagation & 0.913639 $\pm$ 0.0068 & 0.931433 $\pm$ 0.0058 & 0.920626 $\pm$ 0.0146 & 0.939631 $\pm$ 0.0129 & \textbf{0.945314} $\pm$ \textbf{0.0018} \\ 
\cline{2-8}
& \multicolumn{1}{l|}{\multirow{3}{*}{P.cysts imp}} & accuracy & \textbf{0.852084} $\pm$ \textbf{0.0066} & 0.848717 $\pm$ 0.0090 & 0.884709 $\pm$ 0.0152 & \textbf{0.892140} $\pm$ \textbf{0.0144} & \textbf{0.916405} $\pm$ \textbf{0.0074} \\
& \multicolumn{1}{l|}{} & kappa & \textbf{0.755127} $\pm$ \textbf{0.0132} & 0.756045 $\pm$ 0.0138 & 0.811239 $\pm$ 0.0231 & \textbf{0.823333} $\pm$ \textbf{0.0223} & \textbf{0.862764} $\pm$ \textbf{0.0100} \\ 
& \multicolumn{1}{l|}{} & propagation & \textbf{0.845055} $\pm$ \textbf{0.0065} & 0.840924 $\pm$ 0.0056 & 0.880294 $\pm$ 0.0193 & \textbf{0.879200} $\pm$ \textbf{0.0181} & \textbf{0.901996} $\pm$ \textbf{0.0094} \\ 
\cline{2-8}
& \multicolumn{1}{l|}{\multirow{3}{*}{MNIST}} & accuracy & 0.766222 $\pm$ 0.0252 & \textbf{0.852667} $\pm$ \textbf{0.0397} & \textbf{0.901556} $\pm$ \textbf{0.0170} & 0.899778 $\pm$ 0.0096 & \textbf{0.932889} $\pm$ \textbf{0.0136} \\
& \multicolumn{1}{l|}{} & kappa & 0.740028 $\pm$ 0.0280 & \textbf{0.836263} $\pm$ \textbf{0.0440} & \textbf{0.890529} $\pm$ \textbf{0.0119} & 0.888552 $\pm$ 0.0107 & \textbf{0.925403} $\pm$ \textbf{0.0151} \\ 
& \multicolumn{1}{l|}{} & propagation & 0.750571 $\pm$ 0.0320 & \textbf{0.833524} $\pm$ \textbf{0.0362} & \textbf{0.893524} $\pm$ \textbf{0.0064} & 0.895333 $\pm$ 0.0114 & \textbf{0.923619} $\pm$ \textbf{0.0148} \\ 
\cline{2-8}
& \multicolumn{1}{l|}{\multirow{3}{*}{CIFAR10}} & accuracy & 0.228445 $\pm$ 0.0435 & \textbf{0.310000} $\pm$ \textbf{0.0790} & 0.365555 $\pm$ 0.0205 & 0.407778 $\pm$ 0.0136 & 0.424889 $\pm$ 0.0093 \\
& \multicolumn{1}{l|}{} & kappa & 0.142149 $\pm$ 0.0492 & \textbf{0.232875} $\pm$ \textbf{0.0880} & 0.295078 $\pm$ 0.0230 & 0.341907 $\pm$ 0.0148 & 0.360883 $\pm$ 0.0102 \\ 
& \multicolumn{1}{l|}{} & propagation & 0.219048 $\pm$ 0.0428 & \textbf{0.288952} $\pm$ \textbf{0.0790} & 0.356095 $\pm$ 0.0340 & 0.389619 $\pm$ 0.0190 &  0.421143 $\pm$ 0.0126 \\ \hline
\end{tabular}
\end{adjustbox}
\begin{adjustbox}{width=0.9\textwidth}
\begin{tabular}{|c|l|l|c|c|c|c|c|}
\hline
\multirow{21}{*}{\rotatebox[origin=c]{90}{\textbf{(c) DeepFA looping experiment}}} & \multicolumn{1}{l|}{\multirow{3}{*}{H.larvae}} & accuracy & 0.956398 $\pm$ 0.0202 & \textbf{0.974407} $\pm$ \textbf{0.0049} & \textbf{0.974092} $\pm$ \textbf{0.0071} &  \textbf{0.978831} $\pm$ \textbf{0.0074} & \textbf{0.978515} $\pm$ \textbf{0.0031} \\
& \multicolumn{1}{l|}{} & kappa & 0.783412 $\pm$ 0.1150 & \textbf{0.886325} $\pm$ \textbf{0.0299} & \textbf{0.888211} $\pm$ \textbf{0.0298} & \textbf{0.904963} $\pm$ \textbf{0.0353} & \textbf{0.905292} $\pm$ \textbf{0.0138} \\ 
& \multicolumn{1}{l|}{} & propagation & 0.954860 $\pm$ 0.0153 & \textbf{0.976278} $\pm$ \textbf{0.0013} & \textbf{0.970991} $\pm$ \textbf{0.0049} & \textbf{0.981158} $\pm$ \textbf{0.0056} & \textbf{0.980887} $\pm$ \textbf{0.0015} \\ 
\cline{2-8}
& \multicolumn{1}{l|}{\multirow{3}{*}{H.eggs}} & accuracy & \textbf{0.970496} $\pm$ \textbf{0.0039} & \textbf{0.969868} $\pm$ \textbf{0.0086} & \textbf{0.964846} $\pm$ \textbf{0.0237} &  \textbf{0.974262} $\pm$ \textbf{0.0193} & \textbf{0.983679} $\pm$ \textbf{0.0078} \\
& \multicolumn{1}{l|}{} & kappa & \textbf{0.965144} $\pm$ \textbf{0.0046} & \textbf{0.964405} $\pm$ \textbf{0.0102} & \textbf{0.958557} $\pm$ \textbf{0.0278} & \textbf{0.969634} $\pm$ \textbf{0.0227} & \textbf{0.980694} $\pm$ \textbf{0.0093} \\ 
& \multicolumn{1}{l|}{} & propagation & \textbf{0.981676} $\pm$ \textbf{0.0031} & \textbf{0.979251} $\pm$ \textbf{0.0042} & \textbf{0.979520} $\pm$ \textbf{0.0140} & \textbf{0.983563} $\pm$ \textbf{0.0115} & \textbf{0.991107} $\pm$ \textbf{0.0035} \\ 
\cline{2-8}
& \multicolumn{1}{l|}{\multirow{3}{*}{P.cysts}} & accuracy & \textbf{0.889562} $\pm$ \textbf{0.0030} & \textbf{0.925894} $\pm$ \textbf{0.0234} & \textbf{0.938870} $\pm$ \textbf{0.0126} &  \textbf{0.964533} $\pm$ \textbf{0.0120} & \textbf{0.959919} $\pm$ \textbf{0.0088} \\
& \multicolumn{1}{l|}{} & kappa & \textbf{0.853264} $\pm$ \textbf{0.0021} & \textbf{0.900792} $\pm$ \textbf{0.0318} & \textbf{0.918280} $\pm$ \textbf{0.0163} & \textbf{0.952664} $\pm$ \textbf{0.0159} & \textbf{0.946442} $\pm$ \textbf{0.0117} \\
& \multicolumn{1}{l|}{} & propagation & \textbf{0.881800} $\pm$ \textbf{0.0130} & \textbf{0.925321} $\pm$ \textbf{0.0121} & \textbf{0.928783} $\pm$ \textbf{0.0110} & \textbf{0.959570} $\pm$ \textbf{0.0047} & \textbf{0.956726} $\pm$ \textbf{0.0044} \\ 
\cline{2-8}
& \multicolumn{1}{l|}{\multirow{3}{*}{H.eggs imp}} & accuracy & \textbf{0.935680} $\pm$ \textbf{0.0014} & \textbf{0.949370} $\pm$ \textbf{0.0063} & \textbf{0.944372} $\pm$ \textbf{0.0037} & \textbf{0.956758} $\pm$ \textbf{0.0076} & \textbf{0.957844} $\pm$ \textbf{0.0046} \\
& \multicolumn{1}{l|}{} & kappa & \textbf{0.885645} $\pm$ \textbf{0.0033} & \textbf{0.910179} $\pm$ \textbf{0.0111} & \textbf{0.901216} $\pm$ \textbf{0.0078} & \textbf{0.922875} $\pm$ \textbf{0.0130} & \textbf{0.925353} $\pm$ \textbf{0.0080} \\ 
& \multicolumn{1}{l|}{} & propagation & \textbf{0.926681} $\pm$ \textbf{0.0022} & \textbf{0.939352} $\pm$ \textbf{0.0066} & \textbf{0.937675} $\pm$ \textbf{0.0077} & \textbf{0.951369} $\pm$ \textbf{0.0045} & \textbf{0.950252} $\pm$ \textbf{0.0024} \\ 
\cline{2-8}
& \multicolumn{1}{l|}{\multirow{3}{*}{P.cysts imp}} & accuracy & \textbf{0.854522} $\pm$ \textbf{0.0013} & \textbf{0.860908} $\pm$ \textbf{0.0292} & \textbf{0.900035} $\pm$ \textbf{0.0140} & \textbf{0.892488} $\pm$ \textbf{0.0282} & \textbf{0.920933} $\pm$ \textbf{0.0032} \\
& \multicolumn{1}{l|}{} & kappa & \textbf{0.763711} $\pm$ \textbf{0.0026} & \textbf{0.774361} $\pm$ \textbf{0.0434} & \textbf{0.836986} $\pm$ \textbf{0.0217} & \textbf{0.825604} $\pm$ \textbf{0.0450} & \textbf{0.869900} $\pm$ \textbf{0.0051} \\ 
& \multicolumn{1}{l|}{} & propagation & \textbf{0.845652} $\pm$ \textbf{0.0072} & \textbf{0.853915} $\pm$ \textbf{0.0256} & \textbf{0.897367} $\pm$ \textbf{0.0071} & \textbf{0.882684} $\pm$ \textbf{0.0242} & \textbf{0.915335} $\pm$ \textbf{0.0106} \\ 
\cline{2-8}
& \multicolumn{1}{l|}{\multirow{3}{*}{MNIST}} & accuracy & \textbf{0.815778} $\pm$ \textbf{0.0212} & \textbf{0.862222} $\pm$ \textbf{0.0396} & \textbf{0.908444} $\pm$ \textbf{0.0103} & \textbf{0.918000} $\pm$ \textbf{0.0077} & \textbf{0.936444} $\pm$ \textbf{0.0224} \\
& \multicolumn{1}{l|}{} & kappa & \textbf{0.795079} $\pm$ \textbf{0.0236} & \textbf{0.846885} $\pm$ \textbf{0.0439} & \textbf{0.898190} $\pm$ \textbf{0.0114} & \textbf{0.908816} $\pm$ \textbf{0.0086} & \textbf{0.929355} $\pm$ \textbf{0.0249} \\ 
& \multicolumn{1}{l|}{} & propagation & \textbf{0.806000} $\pm$ \textbf{0.0230} & \textbf{0.856667} $\pm$ \textbf{0.0353} & \textbf{0.905905} $\pm$ \textbf{0.0054} & \textbf{0.923238} $\pm$ \textbf{0.0068} & \textbf{0.939905} $\pm$ \textbf{0.0156} \\ 
\cline{2-8}
& \multicolumn{1}{l|}{\multirow{3}{*}{CIFAR10}} & accuracy & \textbf{0.324000} $\pm$ \textbf{0.0418} & \textbf{0.375333} $\pm$ \textbf{0.0436} & \textbf{0.402444} $\pm$ \textbf{0.0125} &  \textbf{0.448445} $\pm$ \textbf{0.0177} & \textbf{0.461555} $\pm$ \textbf{0.0211} \\
& \multicolumn{1}{l|}{} & kappa & \textbf{0.248837} $\pm$ \textbf{0.0463} & \textbf{0.305496} $\pm$ \textbf{0.0483} & \textbf{0.335927} $\pm$ \textbf{0.0138} & \textbf{0.387059} $\pm$ \textbf{0.0199} & \textbf{0.401490} $\pm$ \textbf{0.0236} \\ 
& \multicolumn{1}{l|}{} & propagation & \textbf{0.314286} $\pm$ \textbf{0.0277} & \textbf{0.369334} $\pm$ \textbf{0.0235} & \textbf{0.411238} $\pm$ \textbf{0.0253} & \textbf{0.446667} $\pm$ \textbf{0.0152} & \textbf{0.466857} $\pm$ \textbf{0.0301} \\ \hline
\end{tabular}
\end{adjustbox}
\end{table*}

\noindent\emph{Q2: OPFSemi's effect:} Table~\ref{t.baseline}(b) shows mean and standard deviation for accuracy, Cohen's $\kappa$ coefficient, and propagation accuracy for VGG-16 trained with $S\cup U$, with $U$ labeled by OPFSemi in the 2D projected space for the  \emph{DeepFA} experiment. As for the \emph{baseline} experiment, we see an increase of accuracy and $\kappa$ over the percentages (from $1\%$ to $5\%$) of supervised training samples. The same can be observed for propagation accuracy. Hence, the number of samples correctly assigned by OPFSemi is related to the number of supervised samples used to train VGG-16. Also, the labeling performance of OPFSemi in the 2D projected space can be verified by the propagation accuracy. For the parasites dataset, this accuracy is high (over $80\%$) even when VGG-16 was trained with \emph{just} $1\%$ of the data.\\

\noindent\emph{Q3: Looping OPFSemi effect:} Table~\ref{t.baseline}(c) shows mean and standard deviation of $5$ iterations of \emph{DeepFA looping} for accuracy, Cohen's $\kappa$ coefficient, and propagation accuracy for VGG-16 trained with $S\cup U$, with $U$ labeled by OPFSemi in the 2D projection. As in the \emph{baseline} and \emph{DeepFA} experiments, we see an increase of accuracy and $\kappa$ over the percentages (from $1\%$ to $5\%$) of supervised training samples. The same can be seen for propagation accuracy. Also, the result of $5$ iterations of OPFSemi in the task of labeling samples on the 2D projected feature space can be demonstrated by the propagation accuracy. The propagation accuracy is over $80\%$ even when the VGG-16 feature space was trained with only $1\%$ of data for all the considered datasets, except CIFAR-10.

\vspace{-0.15cm}
\section{Discussion}
\subsection{Added-value of DeepFA looping}
\begin{figure*}[!b]
\centering
\vspace{-0.15cm}
\includegraphics[width=0.87\textwidth]{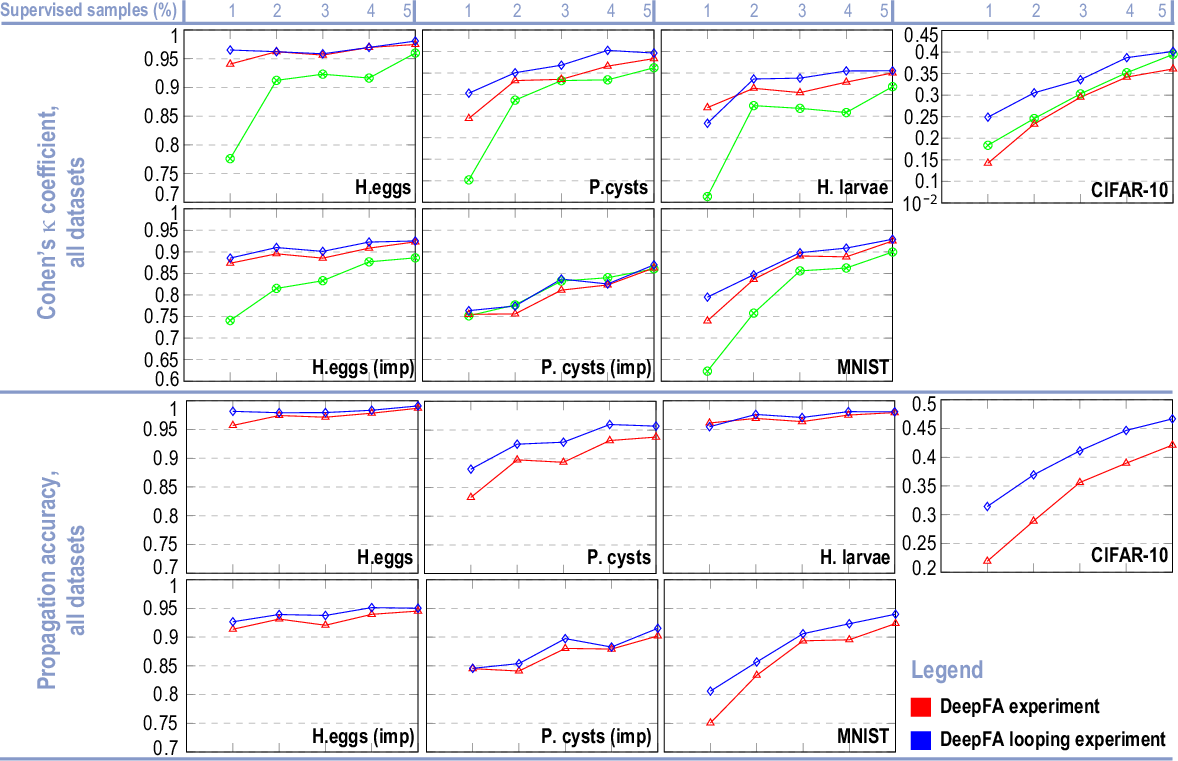}
\vspace{-0.15cm}
\caption{Cohen's $\kappa$ (top) and propagation accuracy (bottom), all datasets, for 1\% to 5\% supervised samples, \emph{DeepFA} (red) \emph{vs} \emph{DeepFA looping} (blue).}
\vspace{-0.15cm}
\label{fig:graphs}
\end{figure*}

Figure~\ref{fig:graphs} plots the average $\kappa$ for our three experiments (\emph{baseline}, \emph{DeepFA}, and \emph{DeepFA looping}), for all 7 studied datasets. We see that \emph{DeepFA looping}  consistently obtains the best results, except for the `P.cysts with impurity' dataset. \emph{DeepFA} shows an improvement over the \emph{baseline} experiment, while the first one was improved by a looping addition in the method.
The obtained gain by \emph{DeepFA looping} is even higher when using a low number of supervised samples. This shows the effectiveness of our method, especially when 
one has very few supervised samples and cannot, or does not want to put effort, into supervising new samples. This gain is lower for CIFAR-10 and almost zero for `P.cysts with impurities', as these datasets are more challenging, as shown by their lowest $\kappa$ scores. 

%ALEX: original graphs. \input{graphs/kappa}

\subsection{Effectiveness of OPFSemi labeling}
The positive results for VGG-16 rely on OPFSemi propagating labels accurately. Figure~\ref{fig:graphs} explains this by showing the average propagation accuracy of OPFSemi for \emph{DeepFA} and \emph{DeepFA looping} for all studied datasets. Propagation accuracy is high for almost all datasets. The hardest dataset, CIFAR-10, has a 50$\%$ propagation accuracy. Yet, for this dataset, the propagation accuracy \emph{gain} given by \emph{DeepFA looping} is  higher than for the other datasets.  We see also the impact of the impurity class for the H.eggs and P.cysts datasets with \emph{vs} without impurity in propagation accuracy (approximately $5\%$). Propagation accuracy is high as long as the sample count increases. The \emph{DeepFA looping} curve is on top of \emph{DeepFA} curve for all datasets, so the effectiveness of OPFSemi label propagation consistently improves by the looping addition. 

%ALEX: original graphs. \input{graphs/propacc}

\subsection{Feature space improvement}
\begin{figure*}[t]
\centering
\vspace{-0.15cm}
\includegraphics[width=1.0\textwidth]{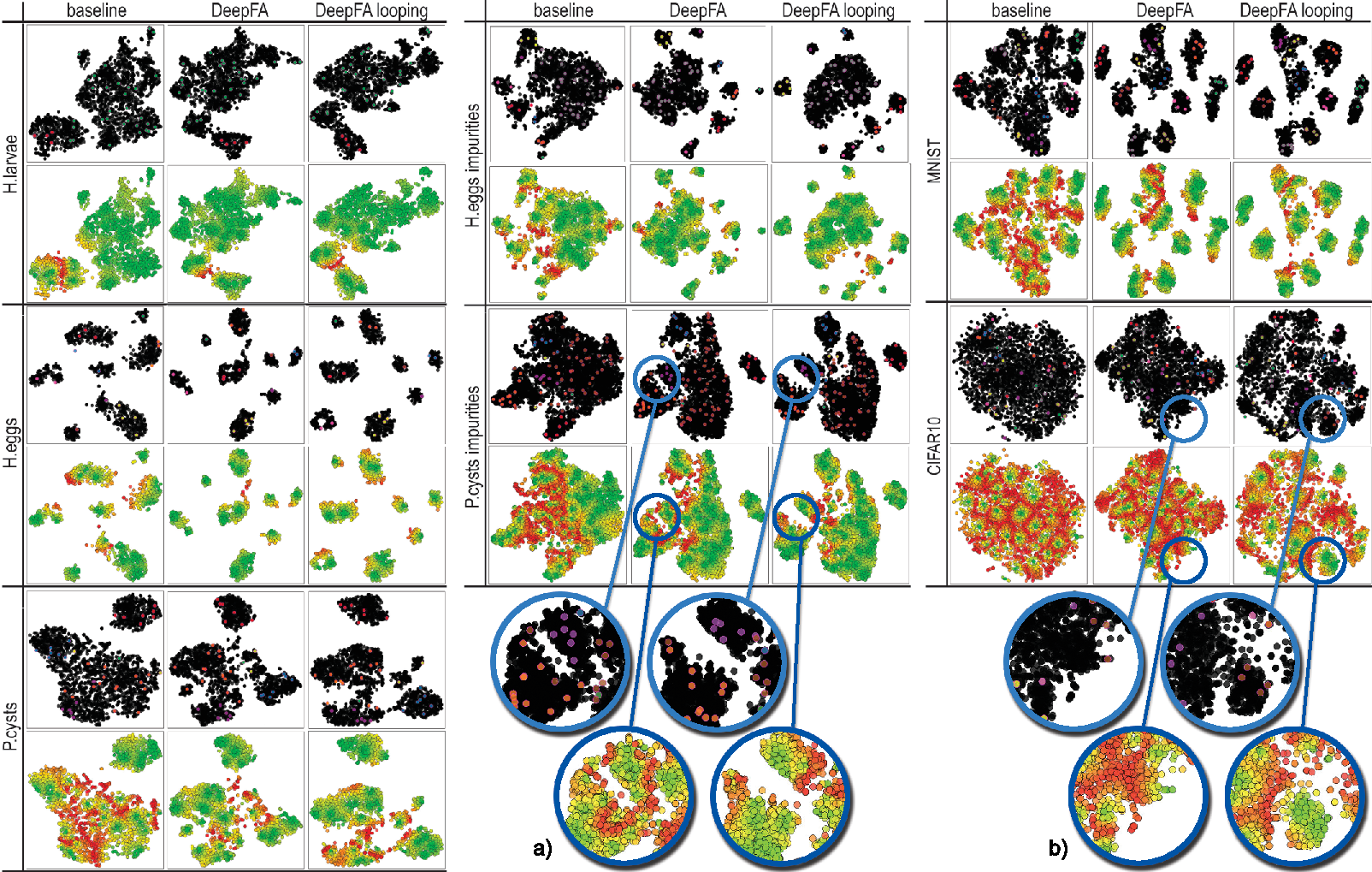}
%\vspace{-0.15cm}
\caption{2D projections of the resulting feature space for the training samples ($S \cup U$) for the \emph{baseline}, \emph{DeepFA}, and \emph{DeepFA looping} experiments, for $x=1\%$ of supervised samples. In the top row for each experiment, supervised samples are colored by true labels, and unsupervised samples are black. In the bottom row of each experiment, samples are colored by OPFSemi's confidence (red=low confidence, green=high confidence) Insets (a,b) show details.}
\vspace{-0.15cm}
\label{fig:images}
\end{figure*}

Figure~\ref{fig:graphs} showed that OPFSemi's label propagation improved VGG-16's effectiveness and also accurately propagated labels for unsupervised samples. Yet, a question remains: Can the OPFSemi labeled samples improve the VGG-16 feature space? Figure~\ref{fig:images} shows this space projected with t-SNE for the studied datasets. Projections are colored by (i) labels (supervised samples colored by the true-label; unsupervised samples black), and (ii) OPFSemi's confidence in classifying a sample (red=low confidence, green=high confidence)~\cite{Miranda:2009,Tavares:2012,Spina:2012}.

For all datasets, we see a significant reduction of red zones from \emph{baseline} to \emph{DeepFA} and a good cluster formation in the projection for same-color (\emph{i.e.}, same-class) supervised samples (Fig.~\ref{fig:images}a). From \emph{DeepFA} to \emph{DeepFA looping}, there is no further reduction of red zones. Yet, we see that different-color groups get more clustered and better separated. We see this clearer for CIFAR-10, which does not show good of cluster separation for the \emph{DeepFA} experiment (Fig.~\ref{fig:images}b). Based on these results, we conclude that OPFSemi's label propagation and the looping strategy improve VGG-16's feature space when this space is fed by those samples.

% Barbara: I did a column version of the same fig to fit in one column, if it is necessary. "projs_iters_column.eps"

\begin{figure*}[bp]
\centering
\vspace{-0.15cm}
\includegraphics[width=\linewidth]{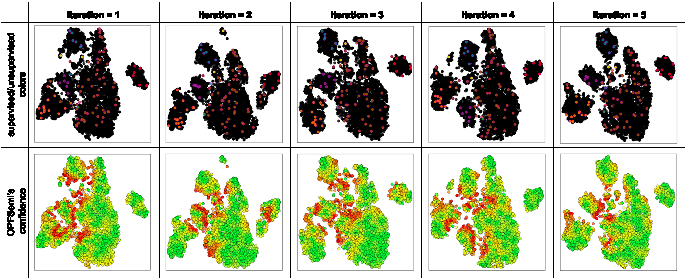}
\vspace{-0.15cm}
\caption{2D projections of training samples ($S\cup U$) for \emph{DeepFA looping} using $1\%$ of supervised samples and the \emph{P. cysts} dataset with impurities. In the top row, projections are colored by class labels, being unsupervised samples shown in black. In the bottom row, projections are colored by the OPFSemi's confidence values (red=low confidence, green=high confidence). Class separation and confidence values increase along with the iterations.}
\label{fig:iterations}.
\end{figure*}

Similarly to Figure~\ref{fig:images}, Figure~\ref{fig:iterations} shows the projected space colored by class labels (unsupervised samples are shown in black) and the OPFSemi's confidence values (from red to green) for $5$ iterations of \emph{DeepFA looping} on  the P. cysts dataset with impurities, using $1\%$ of supervised samples. One can see that class separation and the confidence values increase along with the iterations. It is worth noting that the samples from the red class are well separated from samples of the other classes in the first iteration, but some brown supervised samples get  attached to them in the second iteration. This creates a region of low confidence by OPFSemi in the second iteration. From the third iteration on, the problem is solved.

\subsection{Limitations}
In validating our work, we used only seven datasets, one deep-learning approach (VGG-16), one semi-supervised classifier (OPFSemi), and one projection method (t-SNE). Exploring more (combinations of) such techniques is definitely of extra added value. Also, using more than $5$ looping iterations could help understand how OPFSemi labels low-confidence regions and how it affects the feature space of VGG-16.

\section{Conclusion}
We proposed an approach for increasing the quality of image classification and of extracted feature spaces when lacking large supervised datasets. From a few supervised samples, we create a feature space by a pre-trained VGG-16 model and use the OPFSemi label-propagator to label unsupervised samples on a 2D t-SNE projection of that feature space. We iteratively improve those labels (and the feature space) using labeled samples as input for the VGG-16 training. 

OPFSemi shows low errors when propagating labels and leads VGG-16 to good classification results for several tested datasets. This propagation improves the VGG-16 training and consequently the feature space. The small gain when considering the looping for improvement tells that OPFSemi can stagnate; its label-propagation errors can preclude a better classification result in those cases. This tells that supporting OPFSemi during label propagation would increase VGG-16's classification quality and feature space. To support OPFSemi, we plan next a bootstrapping strategy to avoid propagation in low certainty regions, which can generate misleading labels. We also aim to include user knowledge to support OPFSemi's label propagation, and to understand the VGG-16 training process and feature space generation. This will yield a co-training approach, once a bootstrapping strategy and two classifiers are considered (OPFSemi and VGG-16), and thereby higher quality, and more explainable, deep-learning methods.

%ALEX: We can do this in the final paper
\section*{Acknowledgments}
The authors are grateful to FAPESP grants \#2014/12236-1, \#2019/10705-8, CAPES grants with Finance Code 001, and CNPq grants \#303808/2018-7. The views expressed are those of the authors and do not reflect the official policy or position of the S\~ao Paulo Research Foundation.

\bibliographystyle{elsarticle-num}
\bibliography{refs}

\end{document}